\RequirePackage{amsmath}
\documentclass[runningheads]{llncs}
\usepackage{graphicx} 
\usepackage{amsmath}
\usepackage{ifthen}
\usepackage{adjustbox}
\usepackage{xkeyval}
\usepackage{tikz}
\usepackage{calc}
\usepackage{amssymb}
\usepackage{multirow}
\usepackage{paralist}
\usepackage{comment}
\usepackage[utf8]{inputenc} 
\usepackage[T1]{fontenc}    
\usepackage{url}            
\usepackage{booktabs}       
\usepackage{xcolor}         
\usepackage{arydshln}
\usepackage{wrapfig}
\usepackage{bm}
\usepackage{xspace}
\usepackage{nicefrac}       
\usepackage{microtype}      
\usepackage[hidelinks]{hyperref}
\usepackage{amsmath,amssymb} 
\usepackage[normalem]{ulem}
\useunder{\uline}{\ul}{}
\usepackage{colortbl}
\usepackage[normalem]{ulem}
\usepackage{algorithm}
\usepackage{algorithmic}
\usepackage{arydshln}

\def\eg{\emph{e}.\emph{g}.}
\def\ie{\emph{i}.\emph{e}.}
\newcommand{\norm}[1]{\|#1\|}

\usepackage[capitalize]{cleveref}
\crefname{section}{Sec.}{Secs.}
\Crefname{section}{Section}{Sections}
\Crefname{table}{Table}{Tables}
\crefname{table}{Tab.}{Tabs.}

\newcommand{\mysection}[1]{\vspace{-0.0mm}\section{#1}\vspace{-0.0mm}}
\newcommand{\mysubsection}[1]{\vspace{-0.0mm}\subsection{#1}\vspace{-0.0mm}}

\begin{document}

\title{Multi-View and Multi-Scale Alignment for Contrastive Language-Image Pre-training in Mammography}

\author{
Yuexi~Du\inst{1},~
John~A.~Onofrey\inst{1,2,3},~
Nicha~C.~Dvornek\inst{1,2}~
}
\institute{Departments of 
$^1$Biomedical Engineering, 
$^2$Radiology \& Biomedical Imaging, \\
$^3$Urology, 
Yale University, New Haven, CT, USA \\
}

\authorrunning{Y. Du et al.}

\titlerunning{MaMA}

\maketitle

\begin{abstract}
    Contrastive Language-Image Pre-training (CLIP) demonstrates strong potential in medical image analysis but requires substantial data and computational resources. Due to these restrictions, existing CLIP applications in medical imaging focus mainly on modalities like chest X-rays that have abundant image-report data available, leaving many other important modalities under-explored. 
    Here, we propose one of the first adaptations of the full CLIP model to mammography, which presents significant challenges due to labeled data scarcity, high-resolution images with small regions of interest, and class-wise imbalance.
    We first develop a specialized supervision framework for mammography that leverages its multi-view nature. Furthermore, we design a symmetric local alignment module to better focus on detailed features in high-resolution images. Lastly, we incorporate a parameter-efficient fine-tuning approach for large language models pre-trained with medical knowledge to address data limitations. Our multi-view and multi-scale alignment (MaMA) method outperforms state-of-the-art baselines for three different tasks on two large real-world mammography datasets, EMBED and RSNA-Mammo, with only 52\% model size compared with the largest baseline. 
    \footnote{The code is available at \url{https://github.com/XYPB/MaMA}.}
\end{abstract}

\mysection{Introduction}

Contrastive learning~\cite{chen2020simple,he2019moco,grill2020bootstrap} has become one of the most popular self-supervised representation learning paradigms due to its intuitive concept and robust performance. 
Recently, the introduction of natural language signals to contrastive learning~\cite{radford2021learning} has given rise to modern visual-language models~\cite{li2022blip,li2023blip2,liu2024visual}. Contrastive Language-Image Pre-training (CLIP)~\cite{radford2021learning} has also been applied in the medical imaging domain~\cite{wang2022medclip,huang2021gloria,wang2022multi,zhang2022contrastive,wu2023medklip,zhang2023biomedclip,eslami2023pubmedclip}, showing promising improvement in medical image understanding when large-scale datasets are available~\cite{johnson2019mimic,irvin2019chexpert,eslami2023pubmedclip,zhang2023biomedclip}.
However, the CLIP model in the natural image domain usually demands billion-level image-text pairs to be properly trained~\cite{radford2021learning,sun2023eva,sun2024eva,sun2023evaclip}, which is almost impossible in the medical domain due in part to privacy and security concerns. Existing medical CLIP methods either build general-purpose models with many anatomical sites and modalities from public databases~\cite{eslami2023pubmedclip,zhang2023biomedclip} or focus on single modalities with large-scale (still less than 1M) datasets, \eg, chest X-ray or pathology images~\cite{zhang2022contrastive,huang2021gloria,wang2022multi,wu2023medklip,wang2022medclip,zhou2023advancing,wang2023using,wan2024med,lai2023clipath}. Thus, other imaging modalities, like mammography, have yet to fully benefit from visual-language pre-training.

\begin{figure}[!t]
    \centering
    \includegraphics[width=\columnwidth]{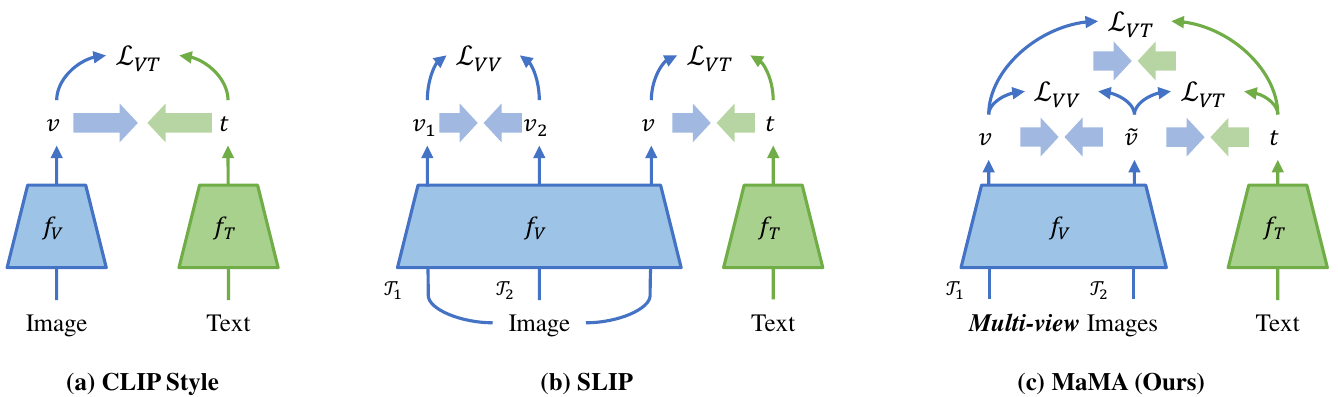}
    \vspace{-5mm}
    \caption{{\bf Comparison of Visual-Language Contrastive Learning Frameworks.} (a) CLIP~\cite{radford2021learning} style; (b) SLIP~\cite{mu2022slip} style; (c) Our proposed MaMA, which aligns image-image and image-text features, exploiting the multi-view nature of mammography.}
    \vspace{-4mm}
    \label{fig:teaser}
\end{figure}  

Mammography is a critical medical imaging modality for screening and diagnosis of breast cancer, one of the most commonly diagnosed cancers globally and a leading cause of cancer-related mortality in women~\cite{sung2021global}. While visual-language pre-training (VLP) has the potential to improve mammography interpretation, there are two major obstacles: 1) \emph{Limited data and annotation}: While recent work has introduced a large-scale mammography image and tabular dataset of more than 110,000 patients~\cite{jeong2023emory}, no corresponding clinical reports are available.
2) \emph{Nature of mammography}: 
Each mammography study usually contains four views of the same patient: left and right side, each with craniocaudal (CC) and mediolateral oblique (MLO) views, giving rise to the critical properties of \emph{bilateral asymmetry}~\cite{donnelly2024asymmirai} and \emph{ipsilateral correspondence}~\cite{liu2021act}. Bilateral asymmetry means images from different sides of the same patient can contain different information, \eg, density, calcification, and mass findings. Ipsilateral correspondence means different views of the same side share similar information from different viewpoints. Clinicians consider both properties and all four images at once when reading a study. Meanwhile, lesions of interest are often relatively small compared with the high-resolution  ($\sim$2,000-by-2,000 pixels) mammograms, which further challenges a model's ability to focus on local details. This pixel-level imbalance compounds the problem of image-level imbalance, in which the vast majority of mammograms will not contain cancer. While recent works~\cite{chen2024mammo,ghosh2024mammo} attempt to address these issues using VLP, they either simply fine-tune a pre-trained CLIP model with a small amount of data~\cite{chen2024mammo} or apply CLIP with same side mammograms and hand-crafted prompts~\cite{ghosh2024mammo}, rather than leveraging the full mammography domain information. 

To address these challenges, we propose a novel \emph{M}ulti-view \emph{a}nd \emph{M}ulti-scale \emph{A}lignment \ie, \emph{MaMA}, contrastive language-image pre-training framework that exploits the multi-view property of mammography and aligns multi-scale features simultaneously. Our work offers the following contributions:
\begin{inparaenum}[\bfseries 1)]
    \item \textbf{Multi-view Design}: We extend the CLIP-style method to leverage the unique multi-view nature of mammography images, introducing (i) inter-study image-to-image contrastive loss, and (ii) symmetric image-text loss.
    \item \textbf{Symmetric Local Alignment (SLA)}: Designed for the relatively small ROIs in mammography, the SLA module improves model understanding of local features without needing dense annotation.
    \item \textbf{Efficient Large Language Model (LLM) as Text Encoder}: Using a parameter-efficient fine-tuned LLM improves understanding of the text while addressing data scarcity.
    \item \textbf{Other Contributions}: We propose two important strategies specifically for mammography VLP: (i) a template-based method to generate structured captions from tabular data that mimics realistic clinical reports and (ii) meta-information masking augmentation to mitigate zero-shot performance loss when training with complex captions.
\end{inparaenum}

We validate our method on two large-scale mammography datasets, EMBED~\cite{jeong2023emory} and RNSA-Mammo~\cite{rsna-breast-cancer-detection}, with multiple settings and compare SOTA medical CLIP methods. The proposed method surpasses all the baselines 
with only 52\% model size, showing promise on multiple mammography-related tasks.

\mysection{Related Works}

\noindent\textbf{Medical Visual-Language Pre-training.} Existing medical VLP approaches can be divided into two types. The first type trains a general-purpose medical CLIP model with large, multi-site, multi-modality datasets derived from PubMed~\cite{eslami2023pubmedclip,zhang2023biomedclip}, scaling dataset size but relying on a vanilla CLIP design~\cite{radford2021learning}. While these models show promise in generalizing across various sites, they can be suboptimal compared to modality-specific models due to the lack of tailored designs for particular imaging modalities. The second type focuses on chest X-ray~\cite{zhang2022contrastive,huang2021gloria,wang2022multi,wu2023medklip,wang2022medclip,zhou2023advancing,wang2023using,wan2024med}, using large datasets like MIMIC-CXR~\cite{johnson2019mimic} or CheXpert~\cite{irvin2019chexpert}, and often relies on single-view images. Some also require full clinical reports~\cite{wang2022multi,wan2024med,zhou2023advancing}, complicating adoption. Recently, a CLIP-based method for mammography was introduced, fine-tuning a pre-trained CLIP model with a multi-view aggregation module~\cite{chen2024mammo}. However, it does not employ contrastive pre-training, does not address pixel-level imbalance, and cannot correlate reports with fine-grained ROIs. Moreover, it only uses a few thousand private cases. Another work proposed Mammo-CLIP \cite{ghosh2024mammo} for pre-training on mammograms but ignored their multi-scale nature and was trained on less than 50k images, limiting generalization and increasing domain shift risk.

\noindent\textbf{Multi-view Contrastive Learning.} Methods like SLIP~\cite{mu2022slip} and DeCLIP~\cite{li2021supervision} combine image-image and image-text loss to achieve robust learning. Similar ideas have been applied to 3D shape recognition~\cite{delitzas2023multi,song2023mv} and action recognition~\cite{shah2023multi}. In mammography~\cite{li2021domain,du2024sift,sun2022breast}, multi-view consistency helps learn high-level shared information. While Mammo-CLIP~\cite{ghosh2024mammo} has attempted to learn intra-side multi-view in a contrastive way, it ignores the correspondence between different sides.

\noindent\textbf{Unsupervised Local Contrastive Learning.} Correlating dense visual representations with fine-grained semantics is crucial for tasks like segmentation. Recent unsupervised methods~\cite{huang2021gloria,wang2022multi,zheng2024dreamlip,wang2023using,liu2023grounding,zhang2024groundhog,shah2023multi,liu2024mlip} have tried to address this challenge. Some rely on pre-trained detectors~\cite{zhang2024groundhog}, others aggregate dense similarity scores into image-level contrastive learning~\cite{zheng2024dreamlip,wang2023using,liu2024mlip,ji2021improving}, while some perform token-level matching with high computational costs~\cite{huang2021gloria,wang2022multi,shah2023multi}. However, none of these learn the explicit local correspondence score for medical images.

\mysection{Method}
\label{sec:method}

\begin{figure}[!t]
    \centering
    \includegraphics[width=\columnwidth]{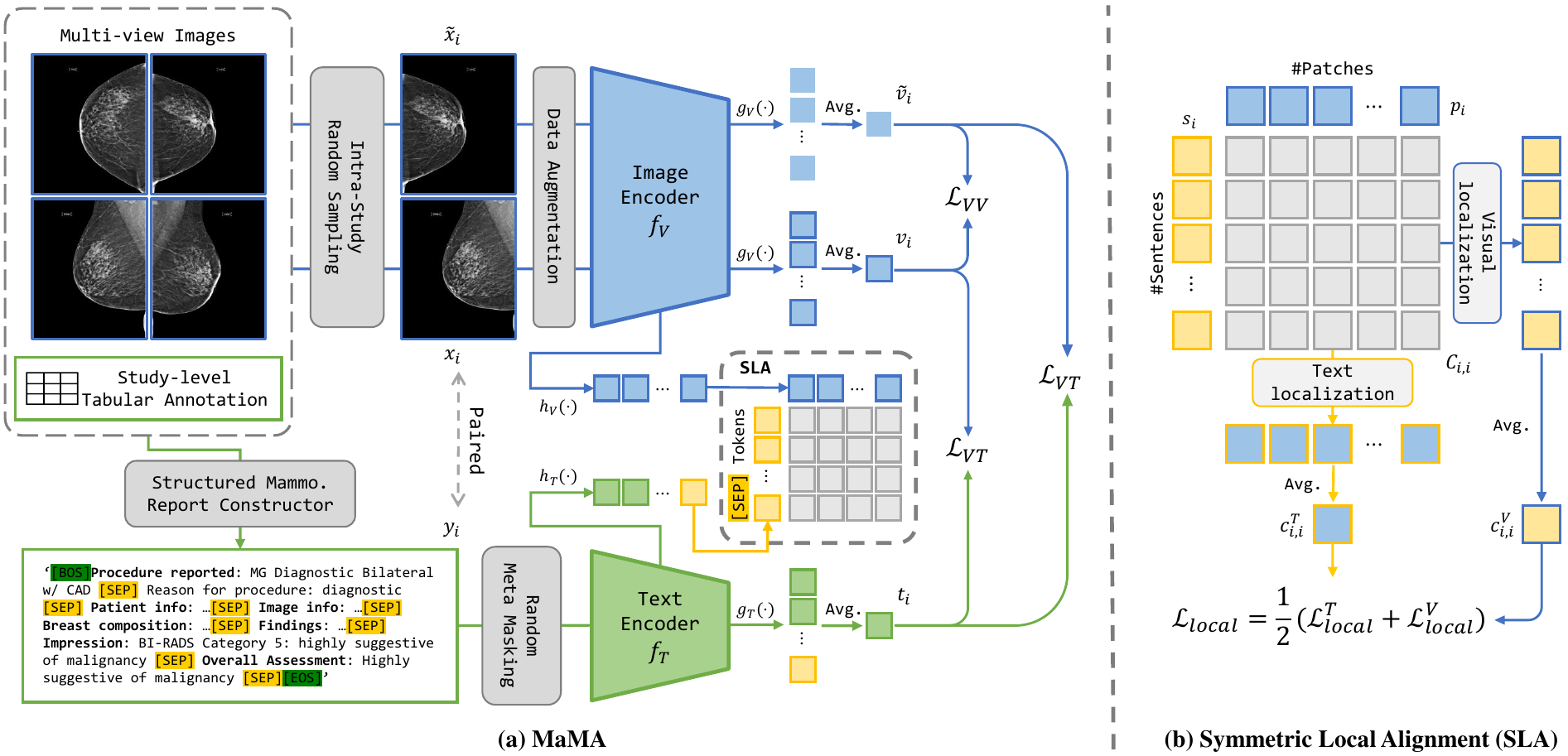}
    \vspace{-4mm}
    \caption{{\bf Proposed Multi-view and Multi-scale (MaMA) VLP Framework.} (a) We utilize the multi-view information of mammography to conduct symmetric image-image and image-text contrastive learning. (b) We localize the most relevant sentence for each image patch and the most relevant patch for each sentence and align these matched local features via symmetric local alignment.}
    \vspace{-2mm}
    \label{fig:method}
\end{figure}

\mysubsection{Structured Report Construction}

Different from datasets that provide paired images with corresponding clinical reports, \eg, MIMIC-CXR~\cite{johnson2019mimic}, large-scale mammography datasets with the full report available are rare. Rather, existing datasets in this domain~\cite{jeong2023emory,rsna-breast-cancer-detection,nguyen2021open} mainly provide a tabular annotation with the anonymized meta-information and the clinical findings, \eg, breast density, calcification findings, and Breast Imaging Reporting and Data System (BI-RADS) assessment category \cite{Sickles2013}. 
Clinical findings serve as cross-validation evidence for the final diagnosis. Using a CLIP-style~\cite{zhao2023clip} caption with only the simple class label will result in a highly simplified caption and limit the model's understanding of the image due to missing details.

We propose a fully automatic template-based caption construction pipeline following the standard clinical report structure~\cite{onken2010digital}~(\cref{fig:method}(a)). We first create a report template with segments describing \textbf{study procedure}, \textbf{patient meta-information}, \textbf{image meta-information}, \textbf{breast composition}, \textbf{findings}, \textbf{clinical impression} and the \textbf{overall assessment} in a clinical report style. Each segment contains keywords that are then replaced with the corresponding information in the tabular data to build a complete clinical report for each image. Our motivation for the structured template is that it can provide more contextual information to the LLM than data in a tabular format. 

\noindent\textbf{Meta-Information Masking.} The increased information from the meta-data may be memorized by the model during pre-training and result in learning shortcuts for the model decision. 
We propose a \textit{data augmentation} method that randomly masks each meta-information keyword with probability $m$ when constructing the caption to force the model to focus on disease-related information.

\mysubsection{Multi-view VLP}

Our multi-view contrastive VLP framework optimizes both image-to-image and symmetric image-to-text contrastive loss (\cref{fig:teaser}(a)).

\noindent\textbf{Multi-view Visual Contrastive Loss} We first optimize the contrastive loss between the multi-view images~(\cref{fig:method}(a)). Let $\mathcal{D}=\{(x_i, y_i),~i=0,1,\dots,N\}$ be a multimodal dataset, where there are $N$ individual images $x_i$ and corresponding text captions $y_i$. We define a study to include the data from the same imaging session for a patient, including one or more image-text pairs. For a random image-text pair $(x_i, y_i) \in \mathcal{D}$, we uniformly sample another image $\tilde{x}_i$ from the same study that $x_i$ belongs to as the positive sample. Note that $\tilde{x}_i$ could be $x_i$ as the augmented view of the same image is naturally a positive sample. We augment both images with random data augmentation and feed into the vision encoder $f_V$ and $d$-dimensional global projection head $g_V$ followed by average pooling to get the visual embedding $v_i,\tilde{v}_i\in\mathbb{R}^{d}$, \ie, $v_i=\mathrm{avg}(g_V(f_V(x_i)))$.
We then compute the similarity for each pair of visual embeddings and optimize the InfoNCE~\cite{chen2020simple} loss for $v_i$ in a mini-batch of size $B$:
\begin{equation}
    \mathcal{L}_{VV}(v_i, \tilde{v}_i) = \log\frac{\exp(sim(v_i,\tilde{v}_i)/\tau_1)}{\sum^B_{j=1}\exp(sim(v_i,v_j)/\tau_1)},~\text{where}~sim(v_i,v_j) = \frac{v_i^T v_j}{\norm{v_i}\norm{v_j}},
\end{equation}
where $\tau_1$ is the visual temperature and $v_j$ is the $j$-th visual embedding in the batch. 
Considering the ipsilateral correspondence, it is natural to treat images on the same side as positive samples of each other, as the features, \eg, tumors, present in one view, are also present in the other view. On the other hand, even if considering bilateral asymmetry for images from different sides, they still share much high-level information such as patient-level features (\eg, breast structure) and similar breast density.
Introducing multi-view mammography contrastive learning forces the model to learn semantically similar features from images within the same study. This also provides a stronger self-supervised signal than using random augmented images. We follow SimCLR~\cite{chen2020simple} to build the image-image contrastive learning framework for simplicity.

\noindent\textbf{Symmetric Visual-Text Contrastive Loss} While existing methods like SLIP~\cite{mu2022slip} also optimize both image-image and image-text contrastive loss, we note there is a potential contradiction between these two objectives when computed for different samples (\cref{fig:teaser}(b)), \ie, $\mathcal{L}_{VV}$ and $\mathcal{L}_{VT}$ are independent and the extra image will increase memory cost. To address this, we propose symmetrically optimizing $\mathcal{L}_{VT}$ and re-using $v_i$.

We feed caption $y_{i}$ to the tokenizer and text encoder $f_T$ and then the text global projection head $g_T$ with average pooling to get the text embedding $t_i\in\mathbb{R}^d$. We optimize the following CLIP~\cite{radford2021learning} loss~(\cref{fig:method}(a)):
\begin{equation}
    \mathcal{L}_{VT}(v_i, t_i) = -\frac{1}{2}(\log\frac{\exp(sim(v_i, t_i)/\tau_2)}{\sum^B_{j=1} \exp(sim(v_i, t_j)/\tau_2)} + \log\frac{\exp(sim(t_i, v_i)/\tau_2)}{\sum^B_{j=1} \exp(sim(t_i, v_j)/\tau_2)}),
\end{equation}
where $\tau_2$ is the learnable CLIP temperature. We symmetrically compute $\mathcal{L}_{VT}$ for both $v_i$ and $\tilde{v}_i$ using the same $t_i$. Namely, we minimize the semantic distance between two views of the same study and the corresponding report simultaneously. Even if the information in $y_i$ is not completely matched with $\tilde{x}_i$, \eg, different side and view information, they still share a large overlap in patient-level information. 
This encourages the model to learn the shared patient-level features via $\mathcal{L}_{VT}(\tilde{v}_i, t_i)$ while focusing on diagnosis-related information by $\mathcal{L}_{VT}(v_i, t_i)$.

\mysubsection{Symmetric Local Alignment (SLA)}
\label{sec:lsa}

Mammography usually contains high-frequency details and the ROI is usually very small. These properties require a higher resolution for the deep learning method to work properly. It also challenges the model's ability to extract important local information and filter out less meaningful background and tissue unrelated to diagnosis. To address these challenges, we propose a symmetric local alignment (SLA) module. Specifically, the SLA module allows the model to determine the local correspondence between each sentence and image patch (\cref{fig:method}(b)). 

We start with extracting local features from input $(x_i, y_i)$. We feed the image and caption to the vision encoder $f_V$ and text encoder $f_T$ respectively, followed by corresponding local projection head $h_V$ and $h_T$ without pooling to produce output feature sequence $v^{local}_i\in\mathbb{R}^{N_V\times d}$ and $t^{local}_i\in\mathbb{R}^{N_T\times d}$, where $N_V$ and $N_T$ are the length of visual tokens and text tokens, respectively. We then extract sentence-level features by selecting the embedding corresponding to the \texttt{[SEP]} token, which results in a sequence of sentence embeddings $s_i\in\mathbb{R}^{S\times d}$, where $S$ is the number of sentences. We extract the image patch-level features by removing the \texttt{[CLS]} tokens, resulting in a sequence of patch embeddings $p_i\in\mathbb{R}^{P\times d}$, where $P$ is the number of patches.
We then compute the sentence-patch correspondence matrix $C_{i,i}\in\mathbb{R}^{S\times P}$ in the form of cosine similarity, which reveals the relevance between local patches and each sentence in the report. However, we cannot directly supervise the learning of this matrix since we have no access to the dense local correspondence. 
Thus, we aggregate the patch-sentence level correspondence matrix $C_{i,i}$ to an image-report level similarity score. 
We start by localizing the patch that has the highest correspondence for each sentence. Namely, we find the most relevant region in the image for each sentence. We call this process \textit{Visual Localization}. We then average the similarity score for each sentence to obtain a correspondence score which describes the similarity of the most relevant patch for the whole report $c_{i,i}^{V}=\frac{1}{S}\sum_j\max_{k}C_{i,i}(j,k)$, where $C_{i,i}(j,k)$ is the similarity between the $j$-th sentence and the $k$-th patch. 
Similarly, we conduct \textit{Text Localization} by finding the most similar sentence for each patch and averaging it to get a score for the similarity of the most relevant sentence for the whole image $c_{i,i}^{T}= \frac{1}{P}\sum_k\max_{j}C_{i,i}(j,k)$. 
We compute the aggregated visual and text local scores for all $p$ and $s$ in the mini-batch and optimize the InfoNCE~\cite{he2019moco} loss:
\begin{equation}
    \mathcal{L}^{V}_{local}(i) = -\frac{1}{2}(\log\frac{\exp(c^V_{i,i}/\tau_{local})}{\sum^B_{j=1}\exp(c^{V}_{i,j}/\tau_{local})} + \log\frac{\exp(c^{V}_{i,i}/\tau_{local})}{\sum^B_{j=1}\exp(c^{V}_{j,i}/\tau_{local})}),
\end{equation}
where $\tau_{local}$ is the local temperature. 
$\mathcal{L}^T_{local}$ is defined similarly.
The final local loss will then be $\mathcal{L}_{local}=\frac{1}{2}(\mathcal{L}^{V}_{local}+\mathcal{L}^T_{local})$. We note that optimizing $\mathcal{L}_{local}$ from the beginning of the training can lead to unstable behavior as the initial visual and language embeddings are noisy. Thus, we add this loss after $\delta$ optimization steps, after the global CLIP loss $\mathcal{L}_{VT}$ is converged.

The intuition behind this design is to mimic the process of radiologic interpretation of a medical image in the real world. On the one hand, in mammography, the clinician will look for the image regions and local features that appear most suspicious for cancer. On the other hand, the clinician will write the radiology report in a few sentences based on the findings across the whole image, while matching each description with a specific feature of the image. Our proposed SLA gives the model the ability to perceive fine-grain local image detail with sentence-level description. The derived local similarity map could also be used as a guide of the relevance between specific image details and each sentence in the provided report and therefore improve the interpretability of the model.

\mysubsection{Overall Pre-training Target} The overall pre-training objective function of our method is given by \cref{eq:overall}:
\begin{equation}
    \label{eq:overall}
    \mathcal{L}(v_i, \tilde v_i, t_i) = \mathcal L_{VV}(v_i, \tilde v_i) + \mathcal L_{VT} (v_i, t_i) + \mathcal L_{VT}(\tilde v_i, t_i) + w\mathcal L_{local}.
\end{equation}
We set $w=0.0$ in the first $\delta=8,000$ training steps and $w=1.0$ afterward.

\mysubsection{LLM with PEFT as Text Encoder}

Lastly, we incorporate parameter-efficient fine-tuning (PEFT) of a pre-trained LLM as our text encoder rather than using the pre-trained BERT encoder~\cite{alsentzer-etal-2019-publicly}. Using a pre-trained LLM with strong domain knowledge can help improve the model's understanding of the text caption and provide a more robust supervised signal for the visual-language pre-training. Moreover, PEFT (\eg, LoRA~\cite{hu2021lora}) can greatly reduce the cost of adapting an LLM when computing resources are limited and maintain strong performance after fine-tuning. Adapting an LLM with PEFT thus has the potential to greatly improve performance while reducing trainable parameters and GPU memory costs compared to learning the commonly adopted BERT-style encoder. To adapt the decoder-only LLM, we use the last non-padding tokens as the caption representation.

\mysection{Experiments}
\label{sec:exp}

\begin{table}[!t]
\centering
\caption{\textbf{BI-RADS Prediction Results on EMBED.} We report balanced accuracy (bACC) and AUC (in \%) of BI-RADS prediction for each method under zero-shot, linear-probing, and full fine-tune settings. We evaluate the effect of training data size on linear probing. $^*$ denotes use of the official pre-trained weights. The best and second-best results are in bold and underlined, respectively. Our method is shaded in gray.}
\label{tab:birads}
\vspace{-2mm}
\setlength{\tabcolsep}{8pt}
\resizebox{\textwidth}{!}
{
\begin{tabular}{lcccccccccc}
\toprule
\multicolumn{1}{c}{\multirow{3}{*}{\textbf{Methods}}} & \multicolumn{2}{c}{\textbf{Zero-shot}} & \multicolumn{6}{c}{\textbf{Linear Probing}} & \multicolumn{2}{c}{\textbf{Full Fine-tune}} \\ \cmidrule(l){2-3} \cmidrule(l){4-9} \cmidrule(l){10-11}
& \multicolumn{2}{c}{\textbf{100\%}} & \multicolumn{2}{c}{\textbf{1\%}} & \multicolumn{2}{c}{\textbf{10\%}} & \multicolumn{2}{c}{\textbf{100\%}} & \multicolumn{2}{c}{\textbf{100\%}} \\
 & bACC & AUC & bACC & AUC & bACC & AUC & bACC & AUC & bACC & AUC \\ \midrule\midrule
\textit{Vision only} &  &  &  &  &  &  &  &  &  &  \\
~~Random-ViT~\cite{dosovitskiy2020image} & - & - & 20.84 & 57.15 & 20.68 & 61.54 & 22.10 & 61.76 & 22.87 & 62.59 \\
~~DiNOv2-ViT~\cite{oquab2023dinov2} & - & - & 22.63 & 61.83 & 25.17 & 66.00 & 29.33 & 70.11 & 30.83 & 71.73 \\ \midrule
\textit{CLIP style} &  &  &  &  &  &  &  &  &  &  \\
~~CLIP~\cite{radford2021learning} & 23.05 & 59.81 & {\ul 26.66} & {\ul 70.35} & {\ul 31.65} & {\ul 74.98} & 34.35 & 74.11 & {\ul 34.25} & 71.61 \\
~~SLIP~\cite{mu2022slip} & 24.14 & {\ul 67.47} & 22.94 & 64.43 & 27.86 & 69.48 & 30.93 & 71.95 & 21.75 & 61.96 \\
~~ConVIRT~\cite{zhang2022contrastive} & 25.27 & 65.13 & 24.62 & 65.09 & 30.38 & 73.33 & 31.27 & 74.03 & 34.54 & {\ul 74.05} \\
~~MGCA~\cite{wang2022multi} & {\ul 26.55} & 63.76 & 23.66 & 64.19 & 30.11 & 72.24 & 30.27 & 72.54 & 34.15 & 73.89 \\
~~MM-MIL~\cite{wang2023using} & 21.78 & 62.41 & 25.85 & 67.16 & 30.94 & 71.99 & {\ul 35.11} & {\ul 76.12} & 33.05 & 71.26 \\
~~Mammo-CLIP-B2$^*$~\cite{ghosh2024mammo} & 16.68 & 56.27 & 23.89 & 62.92 & 23.96 & 66.53 & 24.26 & 66.41 & 25.04 & 69.01 \\
~~Mammo-CLIP-B5$^*$~\cite{ghosh2024mammo} & 16.19 & 55.28 & 23.91 & 64.40 & 24.81 & 69.44 & 24.25 & 69.61 & 25.03 & 70.26 \\
\midrule \rowcolor[HTML]{EFEFEF} 
~~MaMA & \textbf{31.04} & \textbf{74.83} & \textbf{28.46} & \textbf{70.63} & \textbf{35.12} & \textbf{75.98} & \textbf{39.75} & \textbf{77.50} & \textbf{40.31} & \textbf{77.36} \\ \bottomrule
\end{tabular}
}
\vspace{-2mm}

\end{table}

\mysubsection{Pre-training Setup}
\label{sec:pretrain}

\noindent\textbf{Dataset.} We pre-trained our model on the Emory \textbf{EMBED}~\cite{jeong2023emory} dataset, which is one of the largest open mammography datasets. It contains 364,564 2D mammography images for more than $20$k patients collected from 4 hospitals.
The dataset provides tabular annotation about the patient, imaging meta-information, and corresponding study-level findings, \eg, density and BI-RADS. We split it into train/validation/test partitions, each with 70\%/10\%/20\% patients. All the images are resized and padded to $518\times 518$ without changing the aspect ratio to balance the memory cost and batch size requirement.

\noindent\textbf{Implementation Details.} We initialize our image and text encoder with DiNOv2-ViT-B~\cite{oquab2023dinov2} and BioMedLM~\cite{biomedLM2023} respectively, and adapt LoRA~\cite{hu2021lora} to the text encoder for efficient fine-tuning. 
The meta-information masking ratio $m$ is 0.8. We train each model with the AdamW optimizer~\cite{loshchilov2017decoupled} using a learning rate of 4E-5, weight-decay of 0.1, and cosine annealing scheduler for 40k steps. A warm-up for 4k steps is applied. The SLA loss is added after $\delta=8$k steps. We use a batch size of 144 on 4 A5000 GPUs with BFloat-16 precision. Experiments using twice larger image input sizes or batch sizes did not change performance.

\begin{table}[!t]
\centering
\caption{\textbf{Density Prediction Results on EMBED.} We report balanced accuracy (bACC) and AUC (in \%) of density prediction for each method under zero-shot, linear-probing, and full fine-tune settings. We evaluate the effect of training data size on linear probing.  $^*$ denotes use of official pre-trained weights. The best and second-best results are in bold and underlined, respectively. Our method is shaded in gray.}
\label{tab:density}
\vspace{-2mm}
\setlength{\tabcolsep}{8pt}
\resizebox{\textwidth}{!}
{
\begin{tabular}{lcccccccccc}
\toprule
\multicolumn{1}{c}{\multirow{3}{*}{\textbf{Methods}}} & \multicolumn{2}{c}{\textbf{Zero-shot}} & \multicolumn{6}{c}{\textbf{Linear Probing}} & \multicolumn{2}{c}{\textbf{Full Fine-tune}} \\ \cmidrule(l){2-3} \cmidrule(l){4-9} \cmidrule(l){10-11}
\multicolumn{1}{c}{} & \multicolumn{2}{c}{\textbf{100\%}} & \multicolumn{2}{c}{\textbf{1\%}} & \multicolumn{2}{c}{\textbf{10\%}} & \multicolumn{2}{c}{\textbf{100\%}} & \multicolumn{2}{c}{\textbf{100\%}} \\
\multicolumn{1}{c}{} & bACC & AUC & bACC & AUC & bACC & AUC & bACC & AUC & bACC & AUC \\ \midrule\midrule
\textit{Vision only} &  &  &  &  &  &  &  &  &  &  \\
~~Random-ViT~\cite{dosovitskiy2020image} & - & - & 45.81 & 72.83 & 45.11 & 72.62 & 47.01 & 72.92 & 68.47 & 88.61 \\
~~DiNOv2-ViT~\cite{oquab2023dinov2} & - & - & 66.71 & 89.18 & 70.80 & 90.46 & 71.20 & 90.47 & 59.54 & 88.61 \\ \midrule
\textit{CLIP style} &  &  &  &  &  &  &  &  &  &  \\
~~CLIP~\cite{radford2021learning} & 73.56 & {\ul 92.37} & {\ul 74.64} & 91.50 & 75.00 & 90.62 & {\ul 75.97} & 92.39 & 77.47 & \textbf{93.69} \\
~~SLIP~\cite{mu2022slip} & \textbf{75.45} & 92.17 & 73.24 & 91.56 & 74.79 & 92.37 & 75.23 & 92.46 & 64.72 & 86.37 \\
~~ConVIRT~\cite{zhang2022contrastive} & 64.85 & 87.66 & 74.34 & {\ul 92.21} & 74.95 & 92.56 & 74.74 & {\ul 92.58} & {\ul 77.93} & 93.60 \\
~~MGCA~\cite{wang2022multi} & 69.00 & 88.36 & 71.43 & 90.83 & 72.25 & 91.21 & 72.20 & 91.24 & 77.74 & 93.64 \\
~~MM-MIL~\cite{wang2023using} & 69.73 & 89.07 & 74.23 & 91.96 & {\ul 76.69} & {\ul 93.34} & 75.77 & 91.65 & 75.92 & 92.59 \\
~~Mammo-CLIP-B2$^*$~\cite{ghosh2024mammo} & 51.38 & 79.69 & 66.62 & 87.84 & 66.18 & 87.98 & 66.39 & 87.98 & 73.76 & 91.24 \\
~~Mammo-CLIP-B5$^*$~\cite{ghosh2024mammo} & 47.04 & 71.47 & 68.90 & 89.28 & 69.46 & 89.47 & 69.34 & 89.51 & 73.62 & 91.28 \\ \midrule \rowcolor[HTML]{EFEFEF} 
~~MaMA & {\ul 75.40} & \textbf{93.46} & \textbf{76.26} & \textbf{93.11} & \textbf{78.11} & \textbf{93.62} & \textbf{78.09} & \textbf{93.65} & \textbf{78.02} & {\ul 93.65} \\ \bottomrule
\end{tabular}
}
\vspace{-2mm}

\end{table}
\mysubsection{Experimental Setup}
\label{sec:downstream}

\noindent\textbf{Tasks and Datasets.} We primarily evaluate on the \textbf{EMBED}~\cite{jeong2023emory} dataset for both BI-RADS assessment (7 classes) and breast density (4 classes) prediction where the distribution of labels is highly imbalanced. We further sub-sample 7,666 images for BI-RADS prediction and 7,301 images for density prediction from the test split following the dataset distribution for more realistic evaluation. We use all the images with BI-RADS 5 and 6 in the BI-RADS test set to avoid bias due to insufficient data. 
We use the open \textbf{RSNA-Mammo}~\cite{rsna-breast-cancer-detection} dataset with $54$k images as out-of-domain evaluation for binary cancer detection. We split it into train/test sets with 85\%/15\% data.
Given the extremely imbalanced distribution, we report balanced accuracy (bACC) and AUC as our primary metrics. We also report sensitivity and specificity of the RSNA-Mammo cancer detection. 

\noindent\textbf{Evaluation Settings.} We evaluate all methods under zero-shot, linear probing, and full fine-tuning settings. We provide the patient and imaging meta-information in EMBED zero-shot evaluation since it is readily available before the clinician's diagnosis.
For linear probing, we attach a linear classifier and fine-tune it using 1\%, 10\%, or 100\% of the training data following prior work \cite{zhang2022contrastive,huang2021gloria,wang2022multi,wang2022medclip,wu2023medklip,wan2024med}. This data efficiency study with linear probing focuses on the quality of the pre-trained embedding and helps demonstrate the difference between each VLP method. 
For full fine-tuning, we again attach a linear classifier and fine-tune the whole vision model using 100\% of the training data. 
Our learning rate is set to 5E-4 and weight decay to 1E-3 using the SGD optimizer with cosine annealing scheduler for 8k steps with batch size 36. A warm-up of 100 steps is applied.

\noindent\textbf{Baselines.} As one of the first contrastive language-image pre-training methods for mammography, we compare two different styles of baselines: 1) \emph{Vision only}: ViT~\cite{dosovitskiy2020image} models with random initialization and DiNOv2~\cite{oquab2023dinov2} pre-training. 2) \emph{CLIP style}:  \textbf{CLIP}~\cite{radford2019language}; \textbf{SLIP}~\cite{mu2022slip}; \textbf{MM-MIL}~\cite{wang2023using}, which learns local image-language relationships via a multiple instance learning paradigm; \textbf{ConVIRT}~\cite{zhang2022contrastive}, one of the first chest X-ray specific CLIP models; \textbf{MGCA}~\cite{wang2022multi}, which applies multi-granularity feature alignment; and \textbf{Mammo-CLIP}~\cite{ghosh2024mammo}. We pre-train and fine-tune all the CLIP style methods except Mammo-CLIP on our dataset with the same settings and same pre-trained DiNOv2 ViT. For Mammo-CLIP, we use the official pre-trained weights with EfficientNet~\cite{tan2019efficientnet} backbone and the official fine-tuning settings.
We choose not to compare medical VLP methods that adapt domain-specific design and require annotations not present in our dataset~\cite{wang2022medclip,wan2024med,wu2023medklip} or performed worse than our baselines in other studies~\cite{huang2021gloria,zhou2023advancing}. We also do not compare related work that has no released implementation~\cite{liu2024mlip,chen2024mammo}.

\mysubsection{Results}

\begin{table}[!t]
\centering
\caption{\textbf{Cancer Detection Results on RSNA-Mammo~\cite{rsna-breast-cancer-detection}.} We evaluate linear probing and fully fine-tuned settings for the cancer prediction task. We report balanced accuracy (bACC), AUC, sensitivity (SEN), and specificity (SPE) (in \%). $^*$ denotes the use of official pre-trained weights. The best and second-best results are in bold and underlined, respectively. Our method is shaded in gray.}
\label{tab:rsna}
\vspace{-2mm}
\setlength{\tabcolsep}{8pt}
\resizebox{\textwidth}{!}
{
\begin{tabular}{lcccccccccc}
\toprule
\multicolumn{1}{c}{\multirow{2}{*}{\textbf{Methods}}} & \multicolumn{2}{c}{\textbf{Zero-shot}} & \multicolumn{4}{c}{\textbf{Linear Probing}} & \multicolumn{4}{c}{\textbf{Full Fine-tune}} \\ \cmidrule(l){2-3} \cmidrule(l){4-7} \cmidrule(l){8-11}
\multicolumn{1}{c}{} & bACC & AUC & bACC & AUC & SEN & SPE & bACC & AUC & SEN & SPE \\ \midrule\midrule
\textit{Vision only} &  &  &  &  &  &  &  &  &  &  \\
~~Random-ViT~\cite{dosovitskiy2020image} & - & - & 51.90 & 56.34 & {\ul 72.60} & 31.21 & 56.71 & 57.62 & \textbf{77.88} & 35.53 \\
~~DiNOv2-ViT~\cite{oquab2023dinov2} & - & - & 63.23 & 68.59 & 59.62 & 66.84 & 55.12 & 58.18 & {\ul 70.19} & 40.06 \\  \midrule
\textit{CLIP style} &  &  &  &  &  &  &  &  &  &  \\
~~CLIP~\cite{radford2021learning} & 55.74 & 57.85 & 63.89 & 70.28 & 58.17 & {\ul 69.61} & 56.86 & 61.20 & 69.23 & 44.49 \\
~~SLIP~\cite{mu2022slip} & 54.96 & 57.19 & 62.48 & 67.51 & \textbf{78.37} & 46.60 & 56.74 & 60.05 & 63.94 & 49.53 \\
~~ConVIRT~\cite{zhang2022contrastive} & 53.04 & 55.04 & {\ul 65.89} & {\ul 70.70} & 66.83 & 64.96 & 54.53 & 69.85 & 11.06 & \textbf{98.01} \\
~~MGCA~\cite{wang2022multi} & 55.11 & 55.89 & 60.79 & 67.45 & 71.15 & 50.43 & 55.99 & 68.67 & 14.90 & {\ul 97.07} \\ 
~~MM-MIL~\cite{wang2023using} & 53.42 & 53.55 & 64.02 & 70.67 & 58.17 & \textbf{69.86} & 59.97 & 65.04 & 57.21 & 62.73 \\
~~Mammo-CLIP-B2$^*$~\cite{ghosh2024mammo} & {\ul 57.35} & 58.11 & 61.50 & 65.76 & 66.35 & 56.57 & 63.98 & 69.46 & 69.23 & 59.00 \\
~~Mammo-CLIP-B5$^*$~\cite{ghosh2024mammo} & 54.97 & {\ul 59.90} & 62.47 & 67.19 & 62.50 & 61.67 & {\ul 64.57} & {\ul 70.94} & 66.35 & 63.07 \\
\midrule\rowcolor[HTML]{EFEFEF}
~~MaMA & \textbf{60.84} & \textbf{63.55} & \textbf{67.50} & \textbf{73.99} & {\ul 72.60} & 62.40 & \textbf{65.20} & \textbf{73.01} & 67.31 & 63.10 \\ \bottomrule
\end{tabular}
\vspace{-6mm}
}
\end{table}

\noindent\textbf{BI-RADS Prediction.} We report the performance on the EMBED BI-RADS prediction in \cref{tab:birads}. We note that MaMA achieves the best overall performance in all three evaluation settings. Our method outperforms the SoTA baselines by more than 7\% AUC in the zero-shot classification. 
Meanwhile, MaMA also shows the best data efficiency in the linear probing setting. It shows a non-trivial improvement of more than 4\% of balanced accuracy when using full training data. Even when using only 1\% of training data, \eg, less than 10 images for BI-RADS categories 5 and 6, it beats the best baselines by 2\% bACC. 
A similar trend can be found in the full fine-tuning setting, where the gap is more than 6\% bACC.
Mammo-CLIP~\cite{ghosh2024mammo} performs poorly in all 3 settings here, especially in the zero-shot setting. This is likely because the official pre-trained models were pre-trained with much less in-house screening-only mammography data (only have BI-RADS category 0-2), even if using a higher image resolution. 

\noindent\textbf{Density Prediction.} We present the results on EMBED breast density prediction in \cref{tab:density}. Our MaMA model surpassed all the baselines again on most of the metrics. Because the breast density distribution is relatively more balanced, it is reasonable to see a reduced gap between the baselines and our method. Yet, our method still performs either the best or the second best under each evaluation. This demonstrates our model's capability of generalizing on different tasks.

\begin{table}[!t]
\centering
\caption{\textbf{Model Ablation.} Ablation of different designs on the EMBED BI-RADS prediction with balanced accuracy (bACC) and AUC (in \%). The best and second-best results are highlighted in bold and underlined. Our full method is shaded in gray.}
\label{tab:design}
\setlength{\tabcolsep}{8pt}
\resizebox{0.9\textwidth}{!}
{

\begin{tabular}{cccccccccc}
\toprule
\multicolumn{4}{c}{\textbf{Methods}} & \multicolumn{2}{c}{\textbf{Zero-shot}} & \multicolumn{2}{c}{\textbf{Linear Probing}} & \multicolumn{2}{c}{\textbf{Full Fine-tune}} \\ \cmidrule(l){5-6} \cmidrule(l){7-8} \cmidrule(l){9-10} 
SLA & Symm. $\mathcal{L}_{VT}$ & $\mathcal{L}_{VV}$ & PEFT-LLM & bACC & AUC & bACC & AUC & bACC & AUC \\\midrule\midrule
  & \checkmark & \checkmark & \checkmark & 29.28 & 71.16 & 38.71 & {\ul 77.50} & 30.55 & 70.69 \\
 \checkmark &  & \checkmark & \checkmark & {\ul 31.03} & {\ul 72.79} & {\ul 39.57} & 77.39 & {\ul 39.47} & {\ul 76.23} \\
 \checkmark & \checkmark &  & \checkmark & 27.32 & 70.18 & 37.21 & \textbf{77.95} & 23.78 & 63.97 \\
 \checkmark & \checkmark & \checkmark &  & 23.88 & 62.84 & 38.96 & 77.43 & 22.29 & 63.77 \\ \midrule\rowcolor[HTML]{EFEFEF}
 \checkmark & \checkmark & \checkmark & \checkmark & \textbf{31.04} & \textbf{74.83} & \textbf{39.75} & {\ul 77.50} & \textbf{40.31} & \textbf{77.36} \\ \bottomrule
\end{tabular}

}

\end{table}

\begin{table}[t]
  \centering
  \begin{minipage}[t]{0.51\textwidth}
    \centering
    \caption{\textbf{Multi-view ablation.} Different multi-view contrastive strategies.}
    \label{tab:multi}
    \resizebox{\linewidth}{!}
    {
    \setlength{\tabcolsep}{2.0pt}
    \begin{tabular}{lcccccc}
    \toprule
    \multicolumn{1}{c}{\multirow{3}{*}{\textbf{Methods}}} & \multicolumn{6}{c}{\textbf{EMBED BI-RADS}} \\ \cmidrule(l){2-7} 
    \multicolumn{1}{c}{} & \multicolumn{2}{c}{\textbf{Zero-shot}} & \multicolumn{2}{c}{\textbf{Linear Probing}} & \multicolumn{2}{c}{\textbf{Full Fine-tune}} \\
    \multicolumn{1}{c}{} & bACC & AUC & bACC & AUC & bACC & AUC \\ \midrule\midrule
    Same Image & 30.48 & 73.95 & 39.70 & \textbf{77.73} & 39.35 & 76.44 \\
    Intra-side & {\ul 30.71} & {\ul 74.21} & \textbf{39.93} & 77.41 & 35.17 & 76.09 \\
    Intra-study w/o self & 29.33 &  73.21 & 38.20 & 77.49 & {\ul 39.80} & {\ul 76.65} \\ \midrule\rowcolor[HTML]{EFEFEF}
    Intra-study & \textbf{31.04} & \textbf{74.83} & {\ul 39.75} & {\ul 77.50} & \textbf{40.31} & \textbf{77.36} \\ \bottomrule
    \end{tabular}
    }
  \end{minipage}
  \hfill
  \begin{minipage}[t]{0.475\textwidth}
    \caption{\textbf{Caption ablation.} Different text caption construction strategies.}
    \label{tab:caption}
    \centering
    \resizebox{\linewidth}{!}
    {
    \setlength{\tabcolsep}{2.0pt}
    \begin{tabular}{lcccccc}
    \toprule
    \multicolumn{1}{c}{\multirow{3}{*}{\textbf{Methods}}} & \multicolumn{6}{c}{\textbf{EMBED BI-RADS}} \\ \cmidrule(l){2-7} 
    \multicolumn{1}{c}{} & \multicolumn{2}{c}{\textbf{Zero-shot}} & \multicolumn{2}{c}{\textbf{Linear Probing}} & \multicolumn{2}{c}{\textbf{Full Fine-tune}} \\
    \multicolumn{1}{c}{} & bACC & AUC & bACC & AUC & bACC & AUC \\ \midrule\midrule
    CLIP-style & \textbf{35.99} & \textbf{77.66} & 37.74 & 77.25 & 24.00 & 65.35 \\
    Tabular-style & 22.53 & 58.74 & {\ul 38.85} & {\ul 77.40} & {\ul 38.43} & {\ul 76.47} \\
    No Meta Mask & 27.19 & 68.20 & 36.94 & 76.33 & 24.06 & 64.85 \\\midrule\rowcolor[HTML]{EFEFEF}
    Struct. Cap. & {\ul 31.04} & {\ul 74.83} & \textbf{39.75} & \textbf{77.50} & \textbf{40.31} & \textbf{77.36} \\ \bottomrule
    \end{tabular}
    }
  \end{minipage}
  \vspace{-4mm}
\end{table}

\noindent\textbf{Out-of-Domain Data Analysis.} We report performance on the out-of-domain RSNA-Mammo dataset in \cref{tab:rsna}. Since RSNA-Mammo~\cite{rsna-breast-cancer-detection} is an extremely imbalanced dataset (48:1 negative to positive), full fine-tuned models may suffer from overfitting and thus perform worse than linear probing.  
We note our model performs best in terms of balanced accuracy and AUC with a notable gap under all 3 settings. While some baselines outperform our model in either sensitivity or specificity, we note these models are not informative, \ie, they tend to collapse and predict the majority of images to one of the classes. This will lead to a high score in one of the sensitivity or specificity metrics and low performance in the other. 
In contrast, our approach shows reasonable results for both metrics with both sensitivity and specificity greater than 60\%. 
Mammo-CLIP~\cite{ghosh2024mammo} shows a more reasonable performance in this task since it only contains screening mammography and is closer to the pre-training domain of this baseline.

\mysubsection{Ablation Experiments}

\noindent\textbf{Model Design.} We ablate the influence of each component in \cref{tab:design}. We note each component has an important contribution to the overall model performance, as removing any one resulted in inferior performance.
We note that the baseline without PEFT-LLM instead employs BioClinicalBERT~\cite{alsentzer-etal-2019-publicly} and shows a clear drop in zero-shot performance, which validates the importance of using a PEFT-LLM. However, this model still performs well on the linear probing and full fine-tuning tasks, which demonstrates the effectiveness of our other design choices.

\begin{figure}[!t]
    \centering
    \includegraphics[width=0.85\columnwidth]{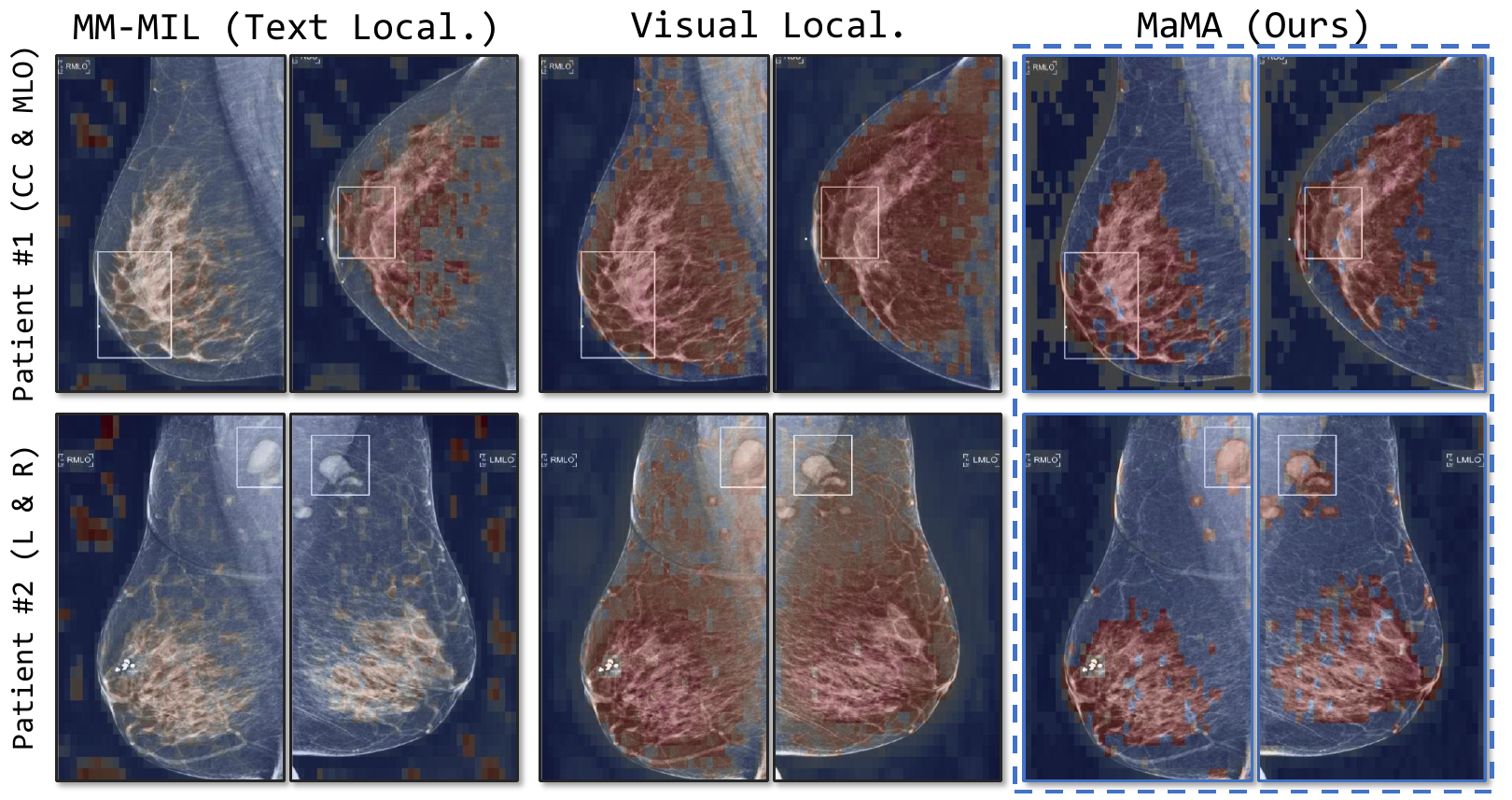}
    \vspace{-2mm}
    \caption{{\bf Local Similarity Maps Overlaid on  Mammograms}. We visualize the learned local similarity map for the ``Impressions'' sentence on test mammograms from  EMBED \cite{jeong2023emory} for MM-MIL~\cite{wang2023using}, our method with only visual localization, and our full method. Heat maps are normalized to [0,1]. The first row shows mammograms from the same side but different view; the second row shows mammograms from the same view but different side. The white boxes denote the dataset-provided annotated ROIs~\cite{jeong2023emory}.}
    \vspace{-4mm}
    \label{fig:local}
\end{figure}  

\noindent\textbf{Multi-view Ablation.} We ablate the multi-view sampling strategy here by using: 1) the same image, 2) an intra-side image, 3) an intra-study image except for the self-augmented view, and 4) all intra-study images and augmented view (\cref{tab:multi}). We note that training with only one image loses the multi-view understanding. Using only intra-side images only considers ipsilateral correspondence and performs worse. Ignoring self-augmented views also degrades performance. 

\noindent\textbf{Caption Ablation.} We evaluate different caption construction strategies in \cref{tab:caption}. We note that a CLIP~style caption that only focuses on class labels shows a better zero-shot performance, but degenerates greatly under linear probing and full fine-tuning. Meanwhile, tabular captions with only key information show the worst zero-shot performance.
Captions without meta-information masking will fail with zero-shot settings since it learned to focus mainly on clinical-irrelevant meta-information during pre-training. Our full design, using a structural caption with meta-information masking augmentation, shows the best performance. 

\noindent\textbf{Local Attention Visualization.} We visualize the learned local patch-sentence similarity map in \cref{fig:local}. We visualize the similarity map for the ``Impression'' sentence, which includes the most important diagnosis information. We note that our methods generally have a better localization quality. The model can accurately locate the high-density and tumor-related regions. The MM-MIL~\cite{wang2023using} with only text localization failed to detect the ROI. Optimizing only visual localization resulted in a vague and inaccurate correspondence map. Our method also shows a multi-view correspondence here compared with the other baselines, especially in the same side image for patient 1.

\mysection{Discussion and Conclusion}
\label{sec:conclusion}
In this work, we presented a novel multi-view and multi-scale alignment contrastive language-image pre-training method for mammography. We proposed utilizing the multi-view nature of mammography and providing local image-sentence correspondence to help address the challenges of small ROIs and high image resolution and provide fine-grained visual clues for decisions. 
The proposed method greatly outperforms multiple existing baselines.
Still, we have yet to evaluate other downstream tasks like object detection and segmentation. 
We will further aim to improve the performance for better clinical usability.

\subsubsection{Acknowledgement} This work was supported by NIH grant R21EB032950.

\bibliographystyle{splncs04}
\bibliography{clip}
\setcounter{tocdepth}{1}

\end{document}